\documentclass[conference]{IEEEtran}
\IEEEoverridecommandlockouts
\usepackage{url}
\usepackage{amsmath}
\usepackage{algorithm}
\usepackage{algorithmic}
\usepackage{graphicx}
\usepackage{array}

\begin{document}
\title{Automatic Graph Partitioning for Very Large-scale Deep Learning
\thanks{\copyright 2021 IEEE. Personal use of this material is permitted.  Permission from IEEE must be obtained for all other uses, in any current or future media, including reprinting/republishing this material for advertising or promotional purposes, creating new collective works, for resale or redistribution to servers or lists, or reuse of any copyrighted component of this work in other works.}}

\author{\IEEEauthorblockN{Masahiro Tanaka\IEEEauthorrefmark{1}, Kenjiro Taura\IEEEauthorrefmark{2}, Toshihiro Hanawa\IEEEauthorrefmark{3}, and Kentaro Torisawa\IEEEauthorrefmark{1}}
\IEEEauthorblockA{\IEEEauthorrefmark{1}Data-driven Intelligent System Research Center (DIRECT)\\
Universal Communication Research Institute, National Institute of Information and Communications Technology (NICT)\\
3-5 Hikaridai, Seika-cho, Soraku-gun, Kyoto 619-0289, Japan\\
Email: \{mtnk, torisawa\}@nict.go.jp}
\IEEEauthorblockA{\IEEEauthorrefmark{2}Department of Information and Communication Engineering\\
Graduate School of Information Science and Technology, University of Tokyo\\
7-3-1 Hongo Bunkyo-ku, Tokyo 113-0033, Japan\\
Email: tau@eidos.ic.i.u-tokyo.ac.jp}
\IEEEauthorblockA{\IEEEauthorrefmark{3}Information Technology Center, University of Tokyo\\
5-1-5 Kashiwanoha, Kashiwa-shi, Chiba, Japan\\
Email: hanawa@cc.u-tokyo.ac.jp}}

\maketitle

\begin{abstract}
This work proposes RaNNC (Rapid Neural Network Connector) as middleware for automatic hybrid parallelism. In recent deep learning research, as exemplified by T5 and GPT-3, the size of neural network models continues to grow. Since such models do not fit into the memory of accelerator devices, they need to be partitioned by model parallelism techniques. Moreover, to accelerate training for huge training data, we need a combination of model and data parallelisms, i.e., hybrid parallelism. Given a model description for PyTorch without any specification for model parallelism, RaNNC automatically partitions the model into a set of subcomponents so that (1) each subcomponent fits a device memory and (2) a high training throughput for pipeline parallelism is achieved by balancing the computation times of the subcomponents. Since the search space for partitioning models can be extremely large, RaNNC partitions a model through the following three phases. First, it identifies atomic subcomponents using simple heuristic rules. Next it groups them into coarser-grained blocks while balancing their computation times. Finally, it uses a novel dynamic programming-based algorithm to efficiently search for combinations of blocks to determine the final partitions. In our experiments, we compared RaNNC with two popular frameworks, Megatron-LM (hybrid parallelism) and GPipe (originally proposed for model parallelism, but a version allowing hybrid parallelism also exists), for training models with increasingly greater numbers of parameters. In the pre-training of enlarged BERT models, RaNNC successfully trained models five times larger than those Megatron-LM could, and RaNNC's training throughputs were comparable to Megatron-LM's when pre-training the same models.
RaNNC also achieved better training throughputs than GPipe on both the enlarged BERT model pre-training (GPipe with hybrid parallelism) and the enlarged ResNet models (GPipe with model parallelism) in all of the settings we tried.
These results are remarkable, since RaNNC automatically partitions models without any modification to their descriptions; Megatron-LM and GPipe require users to manually rewrite the models' descriptions.
\end{abstract}


%

\section{Introduction}

Scaling up deep neural network models has improved performance in a wide variety of tasks.
For example, BERT~\cite{devlin2018bert}, which has up to 340 million parameters, has produced remarkable results in many natural language processing tasks, such as question answering.
Following such successes, the limit on model size has been explored by the latest models, including T5~\cite{raffel2019exploring} (11 billion parameters), Turing-NLG (17 billion parameters), GPT-3~\cite{Brown2020} (175 billion parameters), and GShard~\cite{Lepikhin2020} (600 billion parameters).
Since such enormous models do not fit the memories of accelerator devices (including GPUs), they need to be partitioned by {\em model parallelism} techniques.
Although most frameworks for deep learning support nearly automatic {\em data parallelism} approaches for a wide variety of model architectures, much human effort is still required for model parallelism because model partitioning must be tuned to each model based on a detailed analysis of the computational loads of that model. 

This work proposes RaNNC (Rapid Neural Network Connector), middleware for automatic model partitioning to provide training and inference with very large models. 
Given a model description for PyTorch~\cite{NIPS2019_9015}, which has no specification for model parallelism, RaNNC automatically partitions the model into {\em subcomponents}, each of which is a set of computation tasks such as matrix multiplication.
The subcomponents are determined so that (1) each subcomponent fits into a device memory and (2) high training throughput is achieved for pipeline parallelism~\cite{Giacomoni2008,Gordon2006}.
Furthermore, RaNNC supports {\em hybrid parallelism}, which combines model and data parallelisms, by replicating the partitioned subcomponents to multiple accelerator devices.

Approaches to model partitioning fall into two types: {\em tensor partitioning} and {\em graph partitioning}.
In the former approach, a tensor input to a computation task is partitioned in specified dimensions and computed on different accelerator devices.
The latter partitions a model into subcomponents, each of which is a task graph that consists of atomic operations, including tensor operations, and then distributes the subcomponents to accelerator devices. In general, graph partitioning regards tensor operations as atomic tasks that cannot be further partitioned, while in tensor partitioning, tensor operations are partitioned and computed on multiple accelerator devices.

RaNNC automates graph partitioning. It searches for combinations of computation tasks in the original model.
Since the search space for partitioning a large model can be enormous, RaNNC partitions a model through the following three phases.
First, in the atomic-level partitioning phase, {\em atomic subcomponents} are identified by simple heuristic rules.
Next, in the block-level partitioning phase, the atomic subcomponents are grouped into coarser-grained {\em blocks}.
The search space for model partitioning can be drastically reduced by considering only combinations of coarse-grained blocks as possible subcomponents, rather than all possible combinations of atomic subcomponents.
In this phase, RaNNC attempts to balance the block's computation times so that no particular block becomes a strong bottleneck.
Finally, in the stage-level partitioning phase, a novel dynamic programming-based algorithm efficiently searches for combinations of the blocks 
to determine subcomponents, which are the final results of model partitioning.
Each of the final subcomponents is computed in pipeline parallelism and referred to as a {\em stage}. 
Our search algorithm also determines the number of replicas for each stage to improve the training throughput.

The previous works on automatic graph partitioning, including PipeDream~\cite{Harlap2018}, took a similar approach to RaNNC.
However, they required users to manually specify coarse-grained blocks, while RaNNC automatically determines the blocks so that the search space becomes tractable and no block becomes a strong bottleneck.
In addition, most of the previous works employed asynchronous pipeline parallelism, which suffers from {\em parameter staleness issues}~\cite{Ho2013} as detailed in the next section.
To avoid these issues, RaNNC takes a synchronous approach to pipeline parallelism.

We experimentally compared RaNNC with two popular frameworks, Megatron-LM~\cite{shoeybi2019megatronlm} and GPipe~\cite{huang2018gpipe} (originally proposed for model parallelism, but a version allowing hybrid parallelism also exists).
They take two different approaches to model partitioning: tensor partitioning and graph partitioning.
We pre-trained enlarged BERT models, whose numbers of parameters ranged from 340 million to 12 billion.
The results show that RaNNC successfully trained models five times larger than those trained by Megatron-LM, and its training throughputs were comparable to Megatron-LM's when pre-training the same models.
 RaNNC also achieved better training throughputs than GPipe (hybrid parallelism version) in all of the settings we tried.
For training enlarged ResNet models~\cite{He2016DeepRL} that have up to 3.7 billion parameters, RaNNC also outperformed 
 GPipe (model parallelism version). We could not use Megatron-LM and the hybrid parallelism version of GPipe for ResNet because the former can train only Transformer-based models~\cite{NIPS2017_7181} like BERT and GPT-2, while the available implementation of the latter is applicable only to BERT-based model architectures.
 We also compared RaNNC with PipeDream-2BW, which applies asynchronous pipeline parallelism to increase utilization, and their throughputs were comparable despite PipeDream-2BW suffering from parameter staleness issues.
 
These results are remarkable because RaNNC automatically partitions models without any modification to their descriptions, even though Megatron-LM and GPipe require users to manually rewrite their descriptions so that they fit the special formats needed for model partitioning.

\section{Related Work}

Until very recently, data parallelism has been the dominant approach to distributed deep learning.
However, model parallelism and hybrid parallelism (combining data/model parallelisms) are attracting more and more attention due to the rapid growth of model sizes.
The approaches of the previous works on model partitioning are summarized in Table~\ref{table:partitioning_types}.

\begin{table*}[t]
  \caption{Previous works on model partitioning.}
  \label{table:partitioning_types}
  \begin{tabular}{p{6cm}|p{2cm}|p{1.6cm}|p{1.8cm}|p{1.6cm}|p{2cm}} \hline
     & Partitioning style & Hybrid parallelism & Manual or automatic & Memory estimation & Parameter staleness-free\\ \hline \hline
    Mesh-TensorFlow~\cite{Shazeer2018}, Megatron-LM~\cite{shoeybi2019megatronlm} & Tensor & Yes & Manual & No & Yes \\
    OptCNN~\cite{Jia2018optcnn}, FlexFlow~\cite{Jia2018}, Tofu~\cite{Wang2019} & Tensor & Yes & Auto & No & Yes \\ \hline
    GPipe~\cite{huang2018gpipe} & Graph & No & Manual & No & Yes \\
    AMPNet~\cite{Gaunt2017}, XPipe~\cite{Guan2019} & Graph & No & Manual & No & No \\
    PipeDream~\cite{Harlap2018}, SpecTrain~\cite{Chen2018} & Graph & Yes & Auto & No & No \\
    PipeDream-2BW~\cite{Narayanan2020}, HetPipe~\cite{Park2020} & Graph & Yes & Auto & Yes & No \\ \hline
     RaNNC (Ours) & Graph & Yes & Auto & Yes & Yes \\ \hline
  \end{tabular}
\end{table*}

In this section, we first overview tensor partitioning, which is known as a method for training very large Transformer-based models.
Next, we introduce the previous works on graph partitioning and pipeline parallelism, which are normally combined.
Finally, we describe existing approaches to automatic graph partitioning and address the relevant issues.

\subsection{Tensor partitioning}

Mesh-TensorFlow~\cite{Shazeer2018} and Megatron-LM~\cite{shoeybi2019megatronlm} are frameworks that take a manual approach to tensor partitioning and have successfully trained billion-scale-parameter Transformer-based models~\cite{NIPS2017_7181}, including enlarged BERT and GPT-2.
They require users to hard-code model partitioning using such functions as distributed matrix multiplications.
Since these functions are specialized to Transformer-based model architectures, applying them to other model architectures is difficult.

In existing automatic tensor partitioning approaches~\cite{Jia2018optcnn,Jia2018,Wang2019}, each task type, such as matrix multiplication, is given with several algorithms that optimize the distributed computation of the task type.
Each algorithm has different requirements for input/output data placement on accelerator devices.
According to the required data placement of each task, the previous works automatically choose the appropriate algorithm for each task in a model to reduce the communication overhead.
When a model contains a wide variety of task types, however, it is difficult to implement such algorithms for all of the task types.
In addition, the previous works did not consider memory constraints, which are crucial for training huge models, and they did not report successful training of billion-scale-parameter models.

\subsection{Graph partitioning}

Existing frameworks of manual graph partitioning (including GPipe) require users to partition a model as a task graph into {\em subcomponents}, each of which is also a task graph consisting of such tasks as matrix multiplication and convolution.
In graph partitioning, such tasks are regarded as atomic and cannot be further partitioned.
Unfortunately, when the partitioned subcomponents to be computed on different accelerator devices have sequential dependencies, only one accelerator device can be used at a time.
This severely underutilizes the accelerator devices.

\begin{figure}
\includegraphics[scale=0.45]{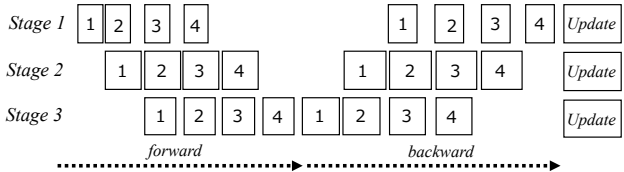}
\caption{Synchronous pipeline parallelism: Numbers in rectangles represent indices of microbatches.}
\label{fig:pipeline}
\end{figure}

Therefore, most works on graph partitioning (including GPipe) employ pipeline parallelism~\cite{BGMZ97,Giacomoni2008,Gordon2006} to increase utilization.
In pipeline parallelism, an accelerator device is allocated to a subcomponent that is often referred to as a {\em stage}.
A mini-batch is split into smaller {\em microbatches} and computed on multiple stages in parallel (Fig.~\ref{fig:pipeline}).
Since the overall computation time is bottlenecked by the stage that takes the longest computation time,
the training throughput is improved by balancing the computation times of the stages.

As mentioned in the previous section, some previous works~\cite{Gaunt2017,Guan2019,Harlap2018,Chen2018,Narayanan2020,Park2020} employed asynchronous pipeline parallelism, which suffers from parameter staleness issues~\cite{Ho2013}.
Such issues are caused by computing a mini-batch using different versions of parameters across stages.
This often results in training that diverges or degrades the quality of learning results due to numerical instability.
Although some works on asynchronous pipeline parallelism proposed various techniques, including weight stashing~\cite{Harlap2018}, to mitigate the negative effects of parameter staleness and showed convergences for relatively small models, other works pointed out that high numerical stability was required for the convergence of huge models~\cite{shoeybi2019megatronlm,Lepikhin2020}.
In fact, no work has reported that asynchronous pipeline parallelism could successfully train billion-scale-parameter models.
Therefore, in this work, we choose a synchronous approach to pipeline parallelism.

\subsection{Automatic graph partitioning}

Since it is often challenging to manually determine stages for pipeline parallelism to achieve a high throughput, some previous works proposed automatic approaches that search for combinations of subcomponents that form stages whose computation times are balanced \cite{Chen2018,Harlap2018,Narayanan2020,Park2020}.

The challenge is that the search space for partitioning a large model can be extremely large.
To reduce the search space, previous works required users to manually specify coarse-grained subcomponents, called {\em blocks} in this work.
For example, PipeDream-2BW~\cite{Narayanan2020} requires users to rewrite a model description to specify which part of the model description code is a {\em layer} to be combined into the final stages.
On the basis of manual preparation, it automatically searches for combinations of the given layers to identify the stages for optimizing estimated training throughput.
Since 192 is the maximum number of layers in their experiments using enlarged BERT models, the search space for partitioning is very small.

\begin{figure*}[t]
\includegraphics[scale=.85]{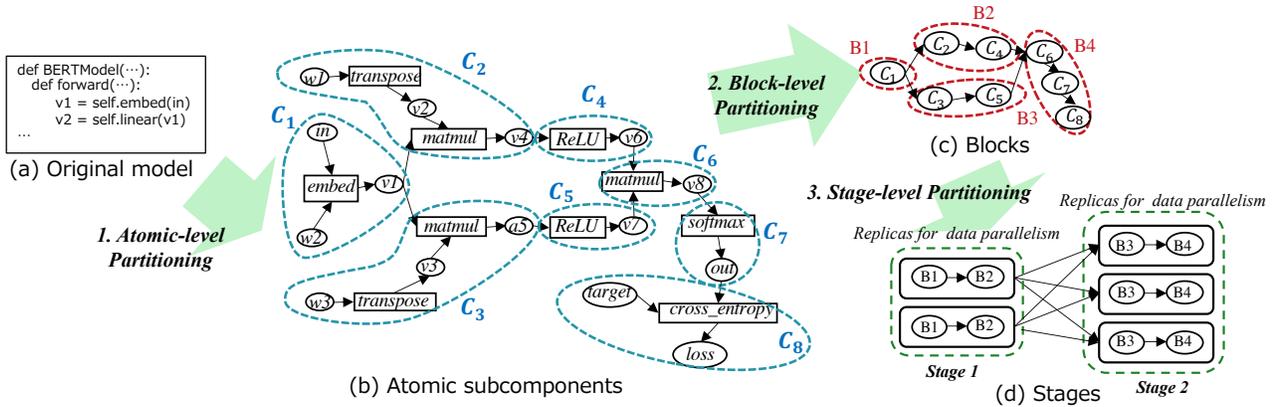}
\vspace{-5mm}
\caption{Partitioning phases.}
\label{fig:partitioning_steps}
\end{figure*}

Such previous works on automatic graph partitioning also assume that the blocks given by users have well-balanced computation times.
If the given blocks are imbalanced, the entire pipeline suffers from a low throughput due to a time-consuming bottleneck stage composed of the blocks.
Although PipeDream-2BW and HetPipe~\cite{Park2020} take layers that repetitively appear as blocks, in their actual models, such intuitively selected layers are not well-balanced.
For example, the last layer of the BERT-Based~\cite{devlin2018bert} model takes 40\% of the overall computation time because it performs matrix multiplication of huge matrices to compute the probabilities of the words in the vocabulary.
This implies that we need to partition the final, coarse-grained layer to finer-grained blocks for improved training throughput.
To address the issue, our work automatically identifies balanced and coarser-grained blocks,
permitting  automatic partitioning of a wide variety of models without human effort.

\section{Partitioning Algorithm}

In this section, we propose RaNNC, which partitions a model to stages and allocates accelerator devices to them.
Its algorithm for model partitioning consists of three phases: {\em atomic-level partitioning}, {\em block-level partitioning}, and {\em stage-level partitioning} (Fig.~\ref{fig:partitioning_steps}).
Starting from the original model, in the atomic-level partitioning phase, we identify {\em atomic subcomponents}.
Next, the block-level partitioning phase groups them into coarser-grained {\em blocks}.
Finally, in the stage-level partitioning phase, we combine the groups into larger subcomponents, which we refer to as {\em stages}. Each stage is allocated an accelerator device, and the entire computation runs in a synchronous pipeline parallelism.

\subsection{Atomic-level partitioning}\label{subsec:atomic_partitioning}

In this phase, we identify {\em atomic subcomponents} that are grouped into stages in the later phases.
We first convert an entire model to a task graph in the manner of the ONNX\footnote{\url{https://onnx.ai/}} format, where there are two types of nodes: {\em tasks} and {\em values}.
The atomic subcomponents are connected subgraphs of the task graph.
An example of a task graph is shown in Fig.~\ref{fig:partitioning_steps}(b), where a rectangle represents a task and a circle represents a value.
The groups of nodes surrounded by blue lines represent atomic subcomponents that are identified by the procedure described below.
To obtain stages whose computation times are well-balanced in our algorithm's final step, we try to keep the atomic subcomponents as fine-grained as possible.

Also, when using hybrid parallelism, combining fine-grained tasks may generate stages that provide only weight parameters or constant values, i.e., values that are independent from the inputs given to the entire model. 
In data parallelism, the replication of such stages is simply a waste of computational resources, so we make sure that each atomic subcomponent has the following property: an atomic subcomponent must contain one {\em non-constant task} whose output changes according to the inputs to the entire model.

Basically, every atomic subcomponent consists of one non-constant task 
and potentially several constant tasks (i.e., tasks that do not depend on the input to the entire model) connected to it.
This guarantees that the output of the subcomponent depends on the input to the entire network, and replicating those atomic subcomponents makes sense.

To create these atomic subcomponents, we use the following procedure, which traverses the task graph twice, once in a forward manner, and once in a backward manner.
First, since non-constant tasks take inputs that are either the input to the entire model or the output of other non-constant tasks, we identify non-constant tasks by exploring a model's task graph from its input in a forward manner, i.e., toward its output. Any task not identified as a non-constant task in this exploration is a constant task.

Then, to create the atomic subcomponents, we traverse the task graph in a backward manner, from the output to the input. Whenever we reach a task classified as a non-constant task, we form a new atomic subcomponent that contains the task and its output value. When we reach a constant task, however, we put it and its output value into the subcomponent to which it is sent.
Note that the output of a constant task can have multiple outgoing edges that target different subcomponents, so in such cases we clone the task and its (constant) predecessors and put each one of them into a target subcomponent.

Using the above simple procedure, we can form atomic subcomponents, each of which has one non-constant task.
For example, in the task graph in Fig~\ref{fig:partitioning_steps}(b), the values $w_1$ and $w_3$ are constant (parameter) values independent from the model's input, and the {\tt transpose} tasks they go through are constant tasks. As such, the {\tt transpose} tasks are connected to the following {\tt matmul} tasks, which are non-constant, to create the atomic subcomponents $C_2$ and $C_3$.

\subsection{Block-level partitioning}

When many atomic subcomponents are identified, the search space of their combinations can still be extremely large.
For example, approximately 15,000 atomic subcomponents are identified in an enlarged BERT model
that has 256 layers.
Therefore, to reduce the search space, we group atomic subcomponents into balanced and coarser-grained subcomponents, which we refer to as {\em blocks}.

For this phase, we introduce two criteria: 1) balance of the blocks' computation times and 2) communication cost.  As mentioned before,
when imbalanced blocks are given, it is difficult to group them into
balanced stages.  In addition, the size of data communicated among stages should be small to reduce the communication overhead.  As shown in 
Fig.~\ref{fig:partitioning_steps}(b), a subcomponent can pass values
to other subcomponents.  Such values are transferred across
accelerator devices when the two subcomponents are placed on different
devices.  Since we group blocks into stages, reducing the size of
values passed between blocks will also reduce the sizes of values
communicated among stages and thus the communication overhead.

To satisfy the above criteria, we use a k-way multilevel partitioning
algorithm, which was previously proposed ~\cite{Karypis1996} and
then extended ~\cite{Huynh2012} for load balancing in streaming applications.
To the best of our knowledge, our work is the first one to use this algorithm for model
partitioning in deep learning.
We adapted the algorithm to our problem as follows.

Our approach consists of the following three steps: {\em coarsening}, {\em uncoarsening}, and {\em compaction}.
The first two steps roughly follow previously proposed procedures ~\cite{Huynh2012}, and the third
is introduced in this work. In the coarsening step, the algorithm
iteratively merges subcomponents to produce coarser-grained
subcomponents. Next, the uncoarsening step attempts to reduce
communication among the subcomponents by adjusting the boundaries between
them.  Then, the compaction step tries to merge subcomponents that
the coarsening step could not merge.  Given a desired number of
blocks $k$ specified by the user, these steps finally
produce $k$ coarse-grained subcomponents, namely blocks.

Let us begin by explaining the coarsening step.
By $A = \{a_i\}^N_{i=1}$, we denote the set of all atomic subcomponents ($a_i$) identified by the procedure described in section \ref{subsec:atomic_partitioning}. 
The coarsening step iteratively forms coarse-grained groups, and we will denote by $G_L$ the set of groups of atomic subcomponents, i.e. blocks, obtained at the $L$-th iteration.
The initial value $G_0$ is a set of singleton sets that contain only one atomic subcomponent and is denoted by $G_0 = \{\{a_i\}\}^N_{i=1}$.

At the $L$-th iteration, the coarsening procedure sequentially picks $v$ in $G_L$ in ascending order of computation time and searches for $w$ in $G_L$ that satisfies the following conditions to create the new group $v \cup w$:

\begin{itemize}
\item $w$ is adjacent to $v$. 
\item The group obtained by merging $v$ and $w$, denoted by $v \cup w$, is {\em convex} and its memory usage is less than the accelerator's device memory size.
\item For any group $w' (\neq w)$ that satisfies the above two conditions, the computation time of $v \cup w$ is smaller than that of $v \cup w'$. 
\end{itemize}

Note that a group $u$ is convex if and only if there is no path
between any pair $\alpha, \beta \in u$ such that the path goes through
any $\gamma \notin u$. A non-convex subcomponent needs to communicate
with other subcomponents to complete its computation. Since stages
are sequentially ordered in pipeline parallelism, a stage that
contains such a subcomponent can cause a deadlock.

\begin{figure}[t]
\centering
\includegraphics[scale=0.38]{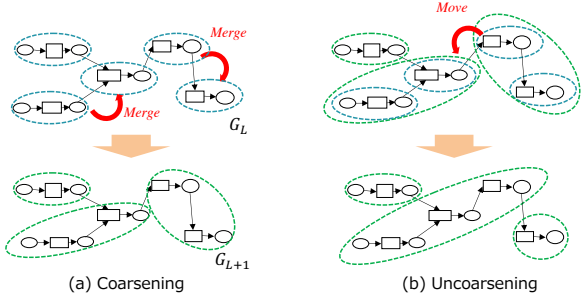}
\vspace{-5mm}
\caption{Block-level partitioning. Dotted circles represent subcomponents. Rectangles and circles represent tasks and values, respectively.}
\label{fig:ml_part}
\end{figure}

If no group satisfies the above conditions, then $v$ is
added to $G_{L+1}$. Otherwise, the new group $v \cup w$ is added to
$G_{L+1}$ (Fig.~\ref{fig:ml_part}(a)).\footnote{To form a new group $v \cup w$, we need to not only take the unions of
subcomponents in $v$ and $w$ but also maintain the edges that connect them in the original task graph; however, we omit such discussion here for the sake of space.}
If there is no more group to be created, then the procedure advances to the $(L+1)$-th iteration and does the same for $G_{L+1}$.
The coarsening step terminates when $|G_L|=k$, or $|G_L|=|G_{L+1}|$, where no group
in $G_L$ is merged, and then the uncoarsening step starts.


The uncoarsening step is used to reduce the communication time between subcomponents.
It starts from processing $G_{L^*}$, which is the set of subcomponent groups obtained in the final iteration of the coarsening step, and proceeds toward $G_0$ iteratively,
decrementing the counter $\tilde{L}$ from $L^*$ to 0. At the
$\tilde{L}$-th step of uncoarsening, for every non-empty group pair
$v$ and $w$ in $G_{\tilde{L}}$ such that $v \cup w \in G_{\tilde{L}+1}$,
the procedure tries to move either $v$ or $w$ into
$t \in G_{\tilde{L}+1}$ such that $t$ is adjacent to $v \cup w$ if the
movement reduces the communication time among subcomponents in the
merged group (Fig.~\ref{fig:ml_part}(b)). 
To measure the improvement, we actually form the groups resulting from the movement and compare the time for communication between the original groups with that between groups resulting from the movement.
We also have to ensure that the resulting groups are convex and fit the device memory.
In addition, the movement of $v$ or $w$ into $t$ is {\em propagated} to $G_{L'}$ for every $L' > \tilde{L}+1$.  For instance, if $v$ is moved from the group containing $v \cup w$ into $t$
at the $\tilde{L}$-th step, $v$ is removed from the group containing $w$
and attached to a group including $t$ in every $G_{L'}$, where $L' > \tilde{L}+1$.
After considering the movement for each group in $G_{\tilde{L}}$, the procedure moves to the next step
$\tilde{L}-1$ and does the same. The uncoarsening step terminates when
$\tilde{L}=0$.

If $|G_{L^*}| = k$, the entire procedure terminates after the uncoarsening step, and $G_{L^*}$ becomes the final result of the block-level partitioning phase.  If $|G_{L^*}| > k$, the final compaction phase
starts. At the beginning, all of the groups in $G_{L^*}$ are topologically
sorted. We can show that, in this sorted list, if group $v$ is
(either left or right) adjacent to a group $w$, $v \cup w$ is a convex
subcomponent. Then, in ascending order of computation times, the
procedure picks up a group $v$ and chooses either the left or right
adjacent group $w$ in the topologically sorted list such that $w$'s
computation time is smaller than that of the alternative (If $w$
is left (or right) adjacent to $v$, the alternative is the right (or
left) adjacent group.). If $v \cup w$ fits the device memory, the
procedure merges them. The procedure repeats this process for every
$v$ in the sorted list until $|G_{L^*}| = k$. When the procedure
terminates, $G_{L^*}$ is the set of final blocks.

In all of the above steps, we need to consider computation times,
memory usages of subcomponents, and time for communication between two
subcomponents. To obtain the computation times and memory usages, we
actually run forward and backward passes of the subcomponents
multiple times and monitor the profiles. The time for communication
between two subcomponents is estimated from the sizes
of the values communicated between the subcomponents and the network
bandwidth of the target environment\footnote{To estimate communication time, we use the intra-node bandwidth, not the inter-node bandwidth. This is because, as described later, our algorithm allocates accelerator devices to stages so that the communication among stages on different nodes is reduced as much as possible.}.

\subsection{Stage-level partitioning}

Once blocks are identified by the block partitioning phase, we search
combinations of the blocks to determine stages as the final results of
our automatic partitioning method. Since the stages can be replicated
for data parallelism, we automatically determine an optimal
number of replicas for each stage at the same time.
The objective in the determination of stages and number of replicas is to minimize the
longest computation time among those of all stages. This is
because, in the synchronous pipeline processing we use, the stage with
the longest computation time becomes the bottleneck and slows down the
speed of the whole computation. 
In our stage-level partitioning phase, we use a dynamic programming algorithm to obtain such stages and numbers of replicas such that the longest computation time is as short as possible. 
We denote a given set of blocks as $B=\{B_1, ..., B_k\}$ and assume that $B$ is topologically sorted according to the dependency of the computation.
Then, when considering $S$ stages, the $i$-th stage is represented as a consecutive sequence of blocks $\{B_j | b_{i-1} < j <= b_{i}\}$, where $b_0 = 0$ and $b_S=|B|$, without losing generality.

We assume that the execution time required for the $i$-th stage includes both the
computation time and the communication time to send the outputs to the
following stage.
The execution time of a stage denoted by a set of blocks $\{B_j | b_{i-1} < j <= b_{i}\}$ depends on the number of the stage's replicas. Thus, we denote the execution time by $h(b_{i-1}, b_i, sd_i)$,
where $sd_i$ is the number of replicas of the stage.

Then, our objective is to minimize the longest execution time of stages $ f_S(\{b_1, \cdots, b_S\}, \{sd_1, \cdots, sd_S\})$, where 
\[
f_S(\{b_1, \cdots, b_S\}, \{sd_1, \cdots, sd_S\})=\max_{1 \leq i \leq S} h(b_{i-1}, b_i, sd_i).
\]

The search algorithm determines $b_i$ and $sd_i$ according to this objective.
We transform this formulation to a recurrence relation for dynamic programming.
We denote the optimal (minimal) value of $f_{\tilde{s}}$ for $\tilde{S}$ stages by $E_{\tilde{S}}$:
\[
E_{\tilde{S}} = \min_{b_{\tilde{S}-1}, d_{\tilde{S}-1}} \max\{E_{\tilde{S}-1}, h(b_{\tilde{S}-1}, b_{\tilde{S}}, d_{\tilde{S}}-d_{{\tilde{S}}-1})\}
\]

Note that the $\tilde{S}$-th stage is denoted by $\{B_j | b_{\tilde{S}-1} < j <= b_{\tilde{S}}\}$, the number of its replicas is denoted by $d_{\tilde{S}}-d_{\tilde{S}-1}$, and $d_{\tilde{S}}$ denotes the total number of replicas for all the stages from the first to the $\tilde{S}$-th.

$E_{\tilde{S}}$ can be efficiently calculated by the dynamic programming-based algorithm shown in Algorithm~\ref{alg:stage_dp}.
In the algorithm, the three outermost loops that iterate on $s$, $b$, and $d$ indicate that we form $s$ stages that contain $b$ blocks and allocate $d$ devices to them in total.
$b'$ and $d'$ used in the following loops represent blocks contained in the previous stages and the number of devices allocated to them.

\begin{algorithm}[H]
\caption{form\_stage\_dp($B$, $S$, $D$, $BS$, $R$, $MB$)}
\label{alg:stage_dp}
\begin{algorithmic}[1]
\STATE $B$: blocks $\{B_1, B_2, ...\}$, $S$: number of stages
\STATE $D$: number of available devices, $BS$: batch size
\STATE $R$: factors of replicas for data parallelism
\STATE $MB$: number of microbatches for pipeline parallelism
\STATE $M$ (constant): size of available device memory
\STATE $d_{min}=1$, $\forall b, d \ V_{s=0,b,d}=0, \ V_{s \neq 0,b,d}=\infty$
\STATE $t_{s=0,b,d}^{f}=0$, $t_{s=0,b,d}^{b}=0$
\FOR{$s$ $\leftarrow$ 1 to $S$}
  \FOR{$b$ $\leftarrow$ $s$ to $|B|$ - $S$ + s}
    \FOR{$d$ $\leftarrow$ $D - (S - s)$ to max($d_{min}$, $s$) {\it (descending)}}
      \FOR{$b'$ $\leftarrow$ $s-1$ to $b-1$}
        \FOR{$d'$ $\leftarrow$ $s-1$ to $d-1$}
           \IF{$V_{s-1,b',d'} = \infty$}
              \STATE continue // Previous stage is infeasible
           \ENDIF
           \STATE $U$ $\leftarrow$ Merge blocks from $B_{b'+1}$ to $B_{b}$
           \STATE $t^{f}$, $t^{b}$, $m$  $\leftarrow$ profile($U$, $\lfloor BS / R / MB / (d-d')\rfloor$)
           \IF{$m > M$}
              \STATE continue // Not fit to device memory
           \ENDIF
           \STATE $\tilde{t}_{s,b,d}^{f} \leftarrow \max \{t_{s-1,b',d'}^{f}, t^{f}\}$
           \STATE $\tilde{t}_{s,b,d}^{b} \leftarrow \max \{t_{s-1,b',d'}^{b}, t^{b}\}$
           \STATE $v$ $\leftarrow$ $\tilde{t}_{s,b,d}^{f} + \tilde{t}_{s,b,d}^{b}$
           \IF{$v < V_{s,b,d}$}
              \STATE $V_{s,b,d} \leftarrow v$
              \STATE $t_{s,b,d}^{f} \leftarrow \tilde{t}_{s,b,d}^{f}, t_{s,b,d}^{b} \leftarrow \tilde{t}_{s,b,d}^{b}$
            \ENDIF
        \ENDFOR
      \ENDFOR
      \IF{No solution with $d$}
        \STATE $d_{min} \leftarrow d + 1$
        \STATE break
      \ENDIF
    \ENDFOR
  \ENDFOR
\ENDFOR
\IF{$V_{S,|B|,D} =\infty$}
   \RETURN INFEASIBLE
\ENDIF
\RETURN ($\{b_1, ..., b_S\}$, $\{d_1 \times R, ..., d_S \times R\}$, $MB$) as a solution \
   where $b_i$ and $d_i$ ($1 \leq i \leq S$) produce $V_{S,|B|,D}$
\end{algorithmic}
\end{algorithm}

In the two inner loops, the algorithm tries to form the $s$-th stage, which contains blocks from $B_{b'+1}$ to $B_b$, using $d - d'$ accelerator devices. 
If there is no solution for $(s-1)$ stages that contains blocks from $B_1$ to $B_{b'}$ using $d'$ accelerator devices, it  proceeds to the next iteration immediately.
Otherwise, it computes a subcomponent that corresponds to the $s$-th stage to profile the execution time and memory requirement.
This profiling is represented by procedure {\tt profile}, which takes a stage to be profiled and a batch size.
Note that we set the batch size to $\lfloor BS / MB / R / (d-d')\rfloor$, where $MB$ and $R$ are the number of microbatches for pipeline parallelism and the factor of the number of replicas, which is described later.
The profiling results include execution times of forward and backward passes, $t^f$ and $t^b$, and required memory $m$.
If $m$ exceeds the available memory, the algorithm moves to the next iteration.
Note that $m$ is the sum of the peak memory usage monitored during forward/backward passes and the memory used for such an optimizer as Adam.
The latter was estimated from the sizes of parameters used in the subcomponents and the type of optimizer.

If the candidate stage $U$ fits the device memory, we compute the longest execution times $\tilde{t}_{s,b,d}^{f}$ and $\tilde{t}_{s,b,d}^{b}$ in all stages.
Then, we evaluate the stages and the allocation of accelerator devices by $v$, the sum of maximum execution times for forward and backward passes, and, if the value is better than the previous best value $V_{s,b,d}$ (the lower value is better), we update $V_{s,b,d}$ and the maximum values of times for forward/backward passes in all stages $t_{s,b,d}^{f}$ and $t_{s,b,d}^{b}$.
Note that we incrementally update the minimum number of accelerator devices, $d_{min}$, to $d+1$ when no solution is found with $d$ devices. Since we do not need to search with $d < d_{min}$, this significantly reduces the search space.

When the algorithm finds a solution, $V_{S,|B|,D}$ is updated by a valid value during the search.
The solution consists of stages as sets of blocks, the number of devices allocated to the stages, and the number of microbatches.

Algorithm~\ref{alg:stage_dp} assumes that the numbers of stages and microbatches are given.
Since we aim to automatically determine those numbers, we also introduce Algorithm~\ref{alg:search} for this purpose.
It first iterates over the number of compute nodes, $n$.
Based on $n$, we determine the number of devices $D$ and the factor of the number of replicas $R$ that are passed to Algorithm~\ref{alg:stage_dp}.
By using a small value of $D$, rather than the total number of accelerator devices, it is possible to reduce the search space in Algorithm~\ref{alg:stage_dp}.
It should be noted that the number of devices $D$ passed to Algorithm~\ref{alg:stage_dp} is aligned to the number of devices of a compute node. 
This also reduces the amount of data communicated across compute nodes.


\begin{algorithm}                      
\caption{form\_stage($N$, $D_{node}$, $BS$)}
\label{alg:search}
\begin{algorithmic}[1]
\STATE $N$: number of nodes, $D_{node}$: number of devices per node
\STATE $B$: blocks $\{B_1, B_2, ...\}$, $BS$: batch size
\STATE $n \leftarrow 1$
\WHILE{$n \leq N$}
  \STATE $D$ $\leftarrow$ $D_{node} \times n$
  \STATE $R$ $\leftarrow$ $N / n$
  \FOR{$S$ $\leftarrow$ ($D_{node} \times (n - 1) + 1$) to $D_{node} \times n$}
    \STATE $A \leftarrow \{\}, MB \leftarrow 1$
    \WHILE{$MB \leq \lfloor B/R \rfloor$}
       \STATE $sol$ $\leftarrow$ form\_stage\_dp($B$, $S$, $D$, $BS$, $R$, $MB$)
       \IF{$sol$ is not INFEASIBLE}
       \STATE Add $sol$ to $A$
       \ENDIF
       \STATE $MB \leftarrow MB \times 2$
    \ENDWHILE
    \IF{$A$ is not empty}
      \RETURN Best $sol$ in $A$
    \ENDIF
  \ENDFOR
  \STATE $n \leftarrow n \times 2$
\ENDWHILE
\RETURN INFEASIBLE
\end{algorithmic}
\end{algorithm}

\section{Experiments}

This section experimentally evaluates RaNNC in terms of training throughputs and scalability regarding the number of model parameters.

\subsection{Experimental setup}

\noindent{\bf Implementation.}
We implemented RaNNC in C++ and python.
In our experiments, RaNNC uses PyTorch 1.6.0 as its backend.
As accelerator devices, RaNNC uses CUDA GPUs.
We used CUDA toolkit 10.2 and cuDNN v7.4.
The inter-process communication is implemented with OpenMPI 4.0.0 and NCCL 2.7.8.

\noindent{\bf Hardware.}
Each compute node in our cluster has two Intel Xeon Gold 6140 processors, 768 GB memory, and eight NVIDIA V100s connected via NVLinks.
Each V100 has 32 GB device memory.
The bandwidth between two V100s is 25 GB/s or 50 GB/s.
The compute nodes are connected by InfiniBand, and the bandwidth is 100 Gbps.

\noindent{\bf Baselines}
We compared RaNNC with three existing frameworks: Megatron-LM, GPipe, and PipeDream-2BW.
All of these frameworks were implemented using PyTorch as their backends.
Megatron-LM uses a type of tensor partitioning that is manually optimized by the framework's developer.\footnote{The version of Megatron-LM used here supports only tensor partitioning, but it was updated to support graph partitioning and pipeline parallelism to train larger models after we conducted the experiments in this section. Note, however, that it still requires a user to manually partition a model to a set of a fixed number of layers.}
The original version of GPipe supports only model parallelism using TensorFlow, but we used two modified versions based on PyTorch.
One of these GPipe versions was developed by the authors of PipeDream-2BW to support hybrid parallelism;
the other, named torchgpipe~\cite{kim2020torchgpipe}, supports model parallelism using only PyTorch.
In this section, we refer to these two versions as GPipe-Hybrid and GPipe-Model, respectively.
PipeDream-2BW automatically partitions a model for hybrid parallelism and employs asynchronous pipeline parallelism, which can cause parameter staleness issues.
In addition to the above frameworks, we also evaluated PyTorch's official implementation as a simple type of data parallelism.

\noindent{\bf Training settings}
We implemented gradient checkpointing~\cite{Griewank2000,Chen2016} for Megatron-LM, GPipe-Hybrid, GPipe-Model, and PipeDream-2BW to reduce the required memory size.
Without gradient checkpointing, during a forward pass, intermediate activations are preserved for the backward pass.
On the other hand, gradient checkpointing discards intermediate activations after a forward pass and recomputes them before running a backward pass.
Consequently, gradient checkpointing increases the computation time but significantly reduces memory usage.
RaNNC automatically implements gradient checkpointing when it partitions a model to more than one stage.

We also used gradient accumulation, which is a popular technique to minimize memory usage, for 
 data parallelism, GPipe-Hybrid, and PipeDream-2BW.
Megatron-LM does not implement gradient accumulation.
In gradient accumulation, a mini-batch is split and computed through a given number of steps while accumulating gradients.
Parameters are updated after processing all of the steps.
The sum of batch sizes that are processed in all steps must be equal to the original batch size.
Since a mini-batch is split into smaller pieces, the memory usage will be reduced.

For RaNNC, we need to set the number of blocks $k$ for the block-level partitioning.
In our experiments, we set $k$ to 32, which we think balances the quality of model partitioning results and the search space for model partitioning.
Since RaNNC allows different numbers of replicas for each stage and actually computes candidate stages, the search space can be very large, even with a small value of $k$, and it takes time to profile the computation time and memory usage for each combination of blocks.

\noindent{\bf Models}
We trained BERT and ResNet models with a larger amount of parameters than originally proposed.
For the enlarged BERT models, we compared RaNNC with data parallelism, Megatron-LM, GPipe-Hybrid, and PipeDream-2BW.
For the enlarged ResNet models, we compared RaNNC with data parallelism and GPipe-Model.
Megatron-LM cannot be applied to ResNet because it is applicable only to Transformer-based models like BERT.
GPipe-Hybrid is also specialized to BERT model architecture, so we could not apply it to ResNet.

To train the enlarged BERT models with RaNNC, we used a model description by NVIDIA\footnote{\url{https://github.com/NVIDIA/DeepLearningExamples/}} without any modification for model partitioning.
The same model description was used for data parallelism.
For Megatron-LM, GPipe-Hybrid, and PipeDream-2BW, we used model descriptions that were optimized by the frameworks' authors.

For the enlarged ResNet models, we used the model description available at PyTorch's official repository for RaNNC and data parallelism.
As for GPipe-Model, we used the model description provided by the framework's authors.

\subsection{Throughputs and scalability}


\noindent{\bf Training of enlarged BERT models}

The original BERT model (BERT-Large), where the hidden layer size is 1024 and the number of layers is 24, has 340 million parameters.
In this experiment, we performed pre-training of enlarged BERT models whose hidden layer sizes were set to 1024, 1536, and 2048, with the number of layers set to 24, 48, 96, 144, 192, and 256. 
The max sequence length was set to 512. 
The largest model we tried in our experiments (256 hidden layers of size 2048) has 12.9 billion parameters.
In all settings, we used 32 GPUs (four servers) with batch size set to 256 (i.e., 8 per GPU).

\begin{figure*}[t]
\centering
\includegraphics[scale=0.9]{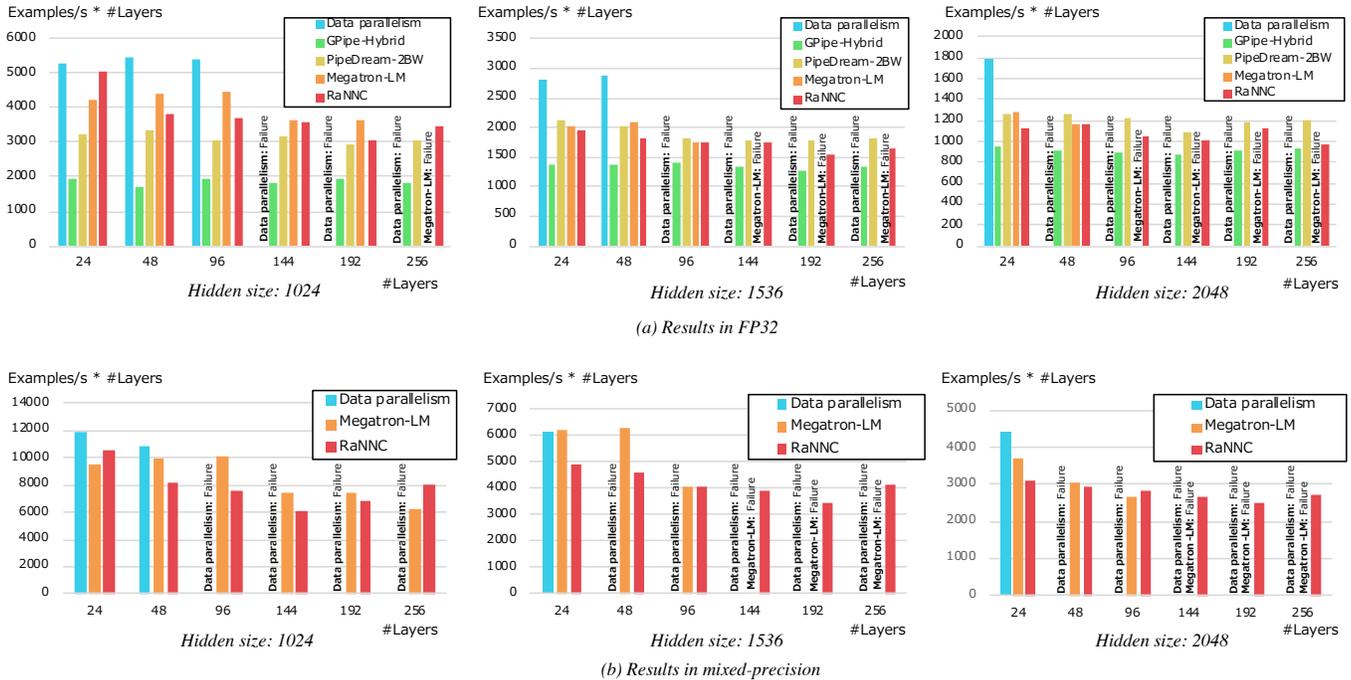}
\caption{Throughputs of enlarged BERT models.}
\label{fig:scale_bert}
\end{figure*}

For GPipe-Hybrid, the user needs to manually determine the number of stages.
As for PipeDream-2BW, although the original paper proposed a method to automatically determine the number of stages, we could not run the available implementation for this purpose.
Therefore, we manually searched for the optimal number of stages for GPipe-Hybrid and PipeDream-2BW.
For these frameworks, the total number of replicas of all stages must match the number of GPUs and the number of layers must be divisible by the number of stages.
In addition, they do not work with a single stage.
Thus, we tried 2, 4, 8, and 16 as the  number of stages and chose the best result.
Megatron-LM also requires users to specify the number of partitions of a given model.
The number must be an exponent of two and be equal to or less than the number of GPUs.
Accordingly, we tried all possible numbers and chose the best result.

Figure~\ref{fig:scale_bert} shows the training throughputs.
We trained models in both FP32 and mixed-precision using Apex AMP\footnote{\url{https://nvidia.github.io/apex}}, which is a de facto standard library for mixed-precision. We trained only RaNNC and Megatron-LM with mixed-precision, since GPipe-Hybrid and PipeDream-2BW do not support it.

The results show that GPipe-Hybrid, PipeDream-2BW, and RaNNC successfully trained all of the models we tried.
The largest model that RaNNC could train was five times larger than those Megatron-LM could train, although Megatron-LM could train larger models than data parallelism.
We attribute this to the fact that it does not implement gradient accumulation.
In addition, the memory efficiency of tensor partitioning was lower than that of graph partitioning in our experiments, since matrix multiplication in tensor partitioning distributes the computational loads, but the size of the buffer to store the results is not reduced.

As for training throughputs in FP32, RaNNC outperformed GPipe-Hybrid in all settings.
We assume this is because RaNNC achieved a better balance of stages than GPipe-Hybrid.
RaNNC can flexibly control computation times of stages by creating different numbers of replicas for each stage, while GPipe-Hybrid creates the same number of replicas for all stages.
On the other hand, as the hidden layer size and the number of layers increase, the differences in throughputs decrease.
This is because, for a very large model, RaNNC needs to partition it to many stages, and the flexibility to determine the number of replicas of stages is restricted.

Since PipeDream-2BW partitions a model in the same way as GPipe-Hybrid, RaNNC can also achieve a better balance of stages than PipeDream-2BW.
PipeDream-2BW slightly outperformed RaNNC in several settings, but it uses asynchronous pipeline parallelism and can cause parameter staleness issues, which can preclude successful training of large-scale models. We believe the throughput gap is tolerable, considering that RaNNC is free of parameter staleness.

As mentioned before, Megatron-LM failed to train large models, but when it succeeded, the throughput was almost the same level as that of RaNNC. These results show that RaNNC is a more effective framework than  Megatron-LM, which requires much human effort for model partitioning.

Finally, to validate the training results obtained using RaNNC, we also performed pre-training of BERT-Large (340 million parameters) and an enlarged BERT model (1.7 billion parameters) using both RaNNC and Megatron-LM.
After the same number of training steps (1 million for BERT-Large and 300k for the enlarged BERT model), we confirmed that RaNNC and Megatron-LM reached almost the same loss value, at 0.176 for BERT-Large and 0.179 for the enlarged BERT model (the difference in loss values resulting from RaNNC and Megatron-LM was less than $1.0 \times 10^{-3}$).

\noindent{\bf Training of enlarged ResNet models}

We also trained enlarged ResNet models to evaluate RaNNC.
The original ResNet has 60 million parameters, but the model was enlarged by various successors.
One of the latest models for image classification, Big Transfer (BiT)~\cite{alex2019big}, adopts a model architecture that multiplies the number of filters of convolutions by certain {\em width factors}.
Following this idea, we also scaled the number of filters and set the width factor to 8.
The largest model used in this experiment (ResNet152x8) has 3.7 billion parameters.

In this experiment, we compared RaNNC with data parallelism and GPipe-Model.
GPipe-Model requires users to partition a model into balanced stages.
Since GPipe-Model can use only GPUs on a single node, the maximum number of stages is eight in our environment.
Since the ResNet model architecture has many more imbalanced layers than BERT and it is difficult to manually optimize the balance of computation times, we tried to partition the models into eight stages in all settings so that the computation times would be as balanced as possible.
We also set the number of microbatches for pipeline parallelism of GPipe-Model to 64.
Recall that RaNNC automatically determines the number of microbatches.
We conducted experiments using four nodes or one node, using GPipe-Model only in the latter setting.
The effective batch size was set to 512 (for 32 GPUs) or 128 (for 8 GPUs).

\begin{figure}[t]
\centering
\includegraphics[scale=0.7]{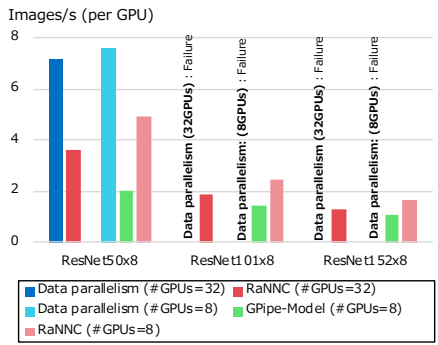}
\caption{Throughputs of enlarged ResNet models. Numbers in model names represent the number of layers and width factors.}
\label{fig:scale_resnet}
\end{figure}

The results in Fig.~\ref{fig:scale_resnet} show that RaNNC and GPipe-Model successfully trained all of the models, while data parallelism could train only the smallest model in our settings.
As for training throughputs, RaNNC outperformed GPipe-Model by a large margin in all of the settings.
We think this is because RaNNC can achieve a better balance of the stages' computation times than can GPipe-Model.
In addition, RaNNC chose an appropriate number of microbatches for each setting, thus increasing utilization.


\subsection{Effect of coarsening}

In our automatic model partition method, the final stages were computed as combinations of relatively coarse-grained blocks, not fine-grained atomic components.
This is for reducing the search space and the costs for estimating computation times and memory consumption for a large number of possible stages.
We evaluated the effectiveness of this method by comparing it with a variant that omits coarsening of atomic components to blocks.
Since we cannot estimate the computation times and memory requirements for all of the stages that should be examined for this variant, we approximated these factors by simply summing those of all atomic subcomponents contained in a stage.

For enlarged BERT models whose hidden layer sizes are 1024, we could train a model with at most 48 layers, and resulting throughputs were 33 percent slower than without using the coarse-grained blocks.
We think this is because estimation by summing computation times of atomic subcomponents results in a considerable overestimation.
Given more layers, the search algorithms in the device allocation step did not finish in 24 hours.

\section{Conclusion}

We proposed RaNNC as middleware for automatic model graph partitioning.
RaNNC automatically partitions a model into a set of subcomponents so that (1) each subcomponent fits into a device memory and (2) high training throughput is achieved for pipeline parallelism.

To evaluate RaNNC, we experimentally compared it with two existing frameworks: Megatron-LM and GPipe.
In pre-training of the enlarged BERT models, RaNNC successfully trained models that are five times larger than those Megatron-LM could train, and RaNNC's training throughputs were comparable to Megatron-LM's when they pre-trained identical models. RaNNC also achieved better training throughputs than GPipe on either the enlarged BERT model's pre-training or the enlarged ResNet models in all of the settings we tried.

In the near future, we plan to evaluate the performances of enormous models that are trained using RaNNC in various applications.





\bibliographystyle{IEEEtran}
\bibliography{rannc_ipdps2021}

\end{document}